\documentclass[letterpaper]{article} 
\usepackage{aaai2026}  
\usepackage{times}  
\usepackage{helvet}  
\usepackage{courier}  
\usepackage[hyphens]{url}  
\usepackage{graphicx} 
\usepackage{natbib}  
\usepackage{caption} 
\usepackage{algorithm}
\usepackage{algorithmic}

\usepackage{newfloat}
\usepackage{listings}
\DeclareCaptionStyle{ruled}{labelfont=normalfont,labelsep=colon,strut=off} 
\floatstyle{ruled}
\newfloat{listing}{tb}{lst}{}
\floatname{listing}{Listing}
\usepackage{url}
\usepackage[utf8]{inputenc} 
\usepackage[T1]{fontenc}    
\usepackage{url}            
\usepackage{booktabs}       
\usepackage{amsfonts}       
\usepackage{nicefrac}       
\usepackage{microtype}      
\usepackage{xcolor}         
\usepackage{amsmath}
\usepackage{amssymb}
\usepackage{mathtools}
\usepackage{amsthm}
\usepackage{epsfig}
\usepackage{amsfonts}
\usepackage{booktabs}
\usepackage{multirow}
\usepackage{float}
\usepackage{amsmath}
\usepackage{amssymb}
\usepackage{bm}
\usepackage{esint}
\usepackage{colortbl}
\usepackage{tabularx}
\usepackage{array} 
\newcolumntype{Y}{>{\centering\arraybackslash}X}
\newlength\savewidth
\newcommand\shline{\noalign{\global\savewidth\arrayrulewidth
		\global\arrayrulewidth 1.1pt}%
	\hline
	\noalign{\global\arrayrulewidth\savewidth}}

\definecolor{navyblue}{rgb}{0.0, 0.0, 0.5}
\definecolor{citecolor}{HTML}{1287AB}
\definecolor{linkcolor}{HTML}{ED1C24}

\title{ASSIST-3D: Adapted Scene Synthesis for Class-Agnostic 3D Instance Segmentation}

\author{
    Shengchao Zhou\textsuperscript{\rm 1},
    Jiehong Lin\textsuperscript{\rm 1}\setcounter{footnote}{1}\thanks{Corresponding authors.},
    Jiahui Liu\textsuperscript{\rm 1},
    Shizhen Zhao\textsuperscript{\rm 1},
    Chirui Chang\textsuperscript{\rm 1} and
    Xiaojuan Qi\textsuperscript{\rm 1}\footnotemark[\value{footnote}]
}
\vspace{5pt}
\affiliations{
    \textsuperscript{\rm 1}The University of Hong Kong\\
    \{sczhou, liujh, zhaosz, xjqi\}@eee.hku.hk, \{mortimer.jh.lin, justinsheunghk\}@gmail.com\\\vspace{5pt}
    \textcolor[HTML]{FF1111}{\url{https://github.com/CVMI-Lab/ASSIST-3D}}
}

\usepackage{bibentry}

\begin{document}
\maketitle

\begin{abstract}
Class-agnostic 3D instance segmentation tackles the challenging task of segmenting all object instances, including previously unseen ones, without semantic class reliance. Current methods struggle with generalization due to the scarce annotated 3D scene data or noisy 2D segmentations. While synthetic data generation offers a promising solution, existing 3D scene synthesis methods fail to simultaneously satisfy geometry diversity, context complexity, and layout reasonability, each essential for this task.
To address these needs, we propose an Adapted 3D Scene Synthesis pipeline for class-agnostic 3D Instance SegmenTation, termed as \textbf{ASSIST-3D}, to synthesize proper data for model generalization enhancement. Specifically, ASSIST-3D features three key innovations, including 
1) \textbf{Heterogeneous Object Selection} from extensive 3D CAD asset collections, incorporating randomness in object sampling to maximize geometric and contextual diversity; 2) \textbf{Scene Layout Generation} through LLM-guided spatial reasoning combined with depth-first search for reasonable object placements; and 3) \textbf{Realistic Point Cloud Construction} via multi-view RGB-D image rendering and fusion from the synthetic scenes, closely mimicking real-world sensor data acquisition.
Experiments on ScanNetV2, ScanNet++, and S3DIS benchmarks demonstrate that models trained with ASSIST-3D-generated data significantly outperform existing methods. Further comparisons underscore the superiority of our purpose-built pipeline over existing 3D scene synthesis approaches.
\end{abstract}


\section{Introduction}
\label{sec:1}

3D instance segmentation tackles the fundamental problem of identifying and delineating  individual object instances in cluttered 3D scenes, with wide applications spanning autonomous driving \cite{guo2024sam,jiang2024mwsis}, robotic navigation \cite{xie2021unseen, yilmaz2024mask4former} and virtual reality \cite{liu2023instance}. This field has been traditionally dominated by class-aware approaches \cite{jiang2020pointgroup,chen2021hierarchical,vu2022softgroup,lu2023query,kolodiazhnyi2024oneformer3d}, which simultaneously segment and classify objects belonging to predefined categories. However, the heavy reliance on annotation restricts these methods to a limited set of object classes, often just dozens, while thousands of diverse, unseen categories exist in real-world scenarios.

\begin{figure*}[t]
\centering
\setlength{\belowcaptionskip}{-5pt}
\includegraphics[width=0.9\linewidth,]{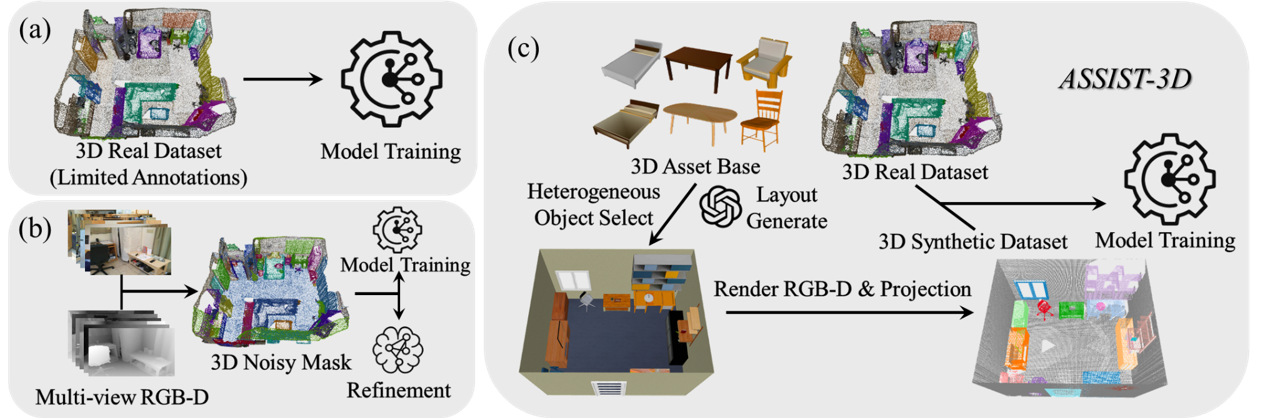}
\caption{(a) Methods adapted from conventional class-aware approaches rely heavily on limited real-world datasets and face severe data scarcity.
(b) Methods that use 2D foundation models for multi-view image segmentation (representing 3D scenes) struggle with 2D segmentation errors or cross-view inconsistency issues.
(c) Our proposed \textbf{ASSIST-3D} addresses these limitations by generating high-quality, fully-annotated synthetic 3D scenes to improve model generalization.
}
\label{img:1}
\end{figure*}

The class-aware bottleneck has driven recent interest in class-agnostic 3D instance segmentation \cite{xu2023sampro3d,huang2024segment3d,rozenberszki2024unscene3d,yangsa3dip}, where the goal shifts to segment all object instances without categorical constraints, including those never encountered during training. Existing solutions primarily explore two strategies, as illustrated in Fig. \ref{img:1} (a) and (b). One kind of approaches \cite{takmaz2023openmask3d,nguyen2024open3dis} modify traditional class-aware 3D architectures by replacing their multi-class classifiers with binary objectness predictors, though their generalization remains hampered by the limited diversity of training objects. Alternatively, methods like those based on 2D foundation models (e.g., SAM \cite{kirillov2023segment}) first perform class-agnostic segmentation on multi-view RGB images, then project 
and fuse the results into 3D according to depth information. While the merged 3D masks can be refined to be final outputs \cite{xu2023sampro3d} or used as pseudo labels for training \cite{huang2024segment3d}, they often suffer from inherent limitations including 2D segmentation inaccuracies and multi-view fusion inconsistencies, ultimately restricting performance and generalization potential.

This work establishes data diversity as the key driver for advancing class-agnostic 3D instance segmentation, where generalization capability directly correlates with the variety and volume of training instances. To address the core challenge of data scarcity, we present \textbf{3D scene synthesis} as a solution. Unlike real-world data collection, which faces costly acquisition, incomplete annotations, and limited scene complexity, synthetic data generation offers a scalable and cost-effective alternative with full annotation control. This paradigm shift enables the creation of large-scale, task-specific 3D datasets with rich supervision, paving the way for generalization. 

To improve data effectiveness, we further identify three fundamental principles for synthetic data, including 1) \textbf{geometry diversity} to ensure broad shape variations, 2) \textbf{context complexity} to capture the variety of objects coexisting in scenes, and 3) \textbf{layout reasonability} to maintain physically plausible global arrangements. Existing synthesis methods \cite{bautista2022gaudi,li2024dreamscene,liu2024pyramid,zhang2024towards}, however, fall short of these criteria, as they either violate the first two principles by biasing towards common real-world objects and their typical neighbors \cite{yang2024holodeck}, or break the third principle through physically implausible random arrangements \cite{rao2021randomrooms}.

To this end, we propose an \textbf{A}dapted \textbf{S}cene \textbf{S}ynthesis pipeline for class-agnostic 3D \textbf{I}nstance \textbf{S}egmen\textbf{T}ation (\textbf{ASSIST-3D}) that satisfies all outlined principles. Specifically, ASSIST-3D comprises three stages, including \textbf{Heterogeneous Object Selection}, \textbf{Scene Layout Generation}, and \textbf{Realistic Point Cloud Construction}. 
First, ASSIST-3D utilizes a large-scale collection of 3D CAD models, e.g. those from Objaverse \cite{deitke2023objaversex,deitke2023objaverse}, as its asset base, and maximizes selection randomness of objects to be placed in one scene to ensure diverse object compositions and contextual arrangements, satisfying the first two principles. 
Second, ASSIST-3D employs large language models for their human-aligned preference to infer plausible spatial relationships and applies a depth-first search strategy for sequential object placement, ensuring layout reasonability. 
Finally, ASSIST-3D mimics real-world point cloud acquisition by rendering multi-view RGB-D images of scenes and merging their partial point clouds to produce realistic point clouds along with instance mask annotations.

We evaluate ASSIST-3D by combining its generated synthetic data with existing real-world datasets to train Mask3D \cite{schult2023mask3d} adapted for class-agnostic segmentation. Experiments on ScanNetV2 \cite{dai2017scannet}, ScanNet++ \cite{yeshwanth2023scannet++}, and S3DIS \cite{armeni20163d} demonstrate consistent performance improvements and enhanced generalization capability with the integration of our synthetic data. The resulting model also significantly outperforms existing state-of-the-art methods.
Furthermore, we conduct comprehensive ablation studies to validate the effectiveness of each individual component, as well as to assess the contribution of the three guiding principles. ASSIST-3D also outperforms existing 3D scene synthesis methods on the target task, including those that overly rely on real-world object selection and neighborhood configurations (e.g., Holodeck \cite{yang2024holodeck}), and those lacking plausible scene layouts (e.g., RandomRooms \cite{rao2021randomrooms}), both of which fall short of fully adhering to the proposed principles. In summary, our technical contributions are three-fold:

\begin{itemize}
 \item We repurpose 3D scene synthesis as a solution to the data scarcity challenge in class-agnostic 3D instance segmentation and introduce ASSIST-3D, a novel, purpose-built pipeline for generating fully annotated synthetic 3D scene data to enhance model generalization.
\item ASSIST-3D consists of three stages, including Heterogeneous Object Selection, Scene Layout Generation, and Realistic Point Cloud Construction, to ensure geometry diversity, context complexity, and layout reasonability.
\item Experiments on the ScanNetV2, ScanNet++, and S3DIS benchmarks demonstrate the effectiveness of the synthetic data generated by ASSIST-3D, with the resulting model significantly outperforming existing methods. Additional comparisons highlight the advantages of our purpose-built pipeline over existing 3D scene synthesis approaches.
\end{itemize}

\section{Related Work}
\label{sec:2}
\paragraph{Class-agnostic 3D Instance Segmentation}
Class-agnostic 3D instance segmentation advances beyond the traditional class-aware setting by removing category dependencies, enabling segmentation of arbitrary objects in 3D scenes, including objects from categories not seen during training.

Current approaches address the generalization challenge through two main strategies. The first strategy \cite{takmaz2023openmask3d,yan2024maskclustering,nguyen2024open3dis,boudjoghra2024open} modifies conventional class-aware 3D architectures by eliminating category dependencies and instead learning the concept of "objectness" to differentiate object segments from non-object regions. For instance, OpenMask3D \cite{takmaz2023openmask3d} adapts the framework of Mask3D by replacing its multi-class classifier with a binary objectness predictor. While these methods leverage existing annotated 3D scene datasets for training, their generalization capability remains constrained by the limited quantity of annotated objects available in current datasets.
The second strategy \cite{xu2023sampro3d,huang2024segment3d,rozenberszki2024unscene3d,yin2024sai3d,yangsa3dip} capitalizes on the strong generalization performance of 2D foundation models, which benefit from large-scale training data and have demonstrated robust performance across diverse downstream tasks. These methods typically employ 2D foundation models to segment objects in multi-view rendered images of a 3D scene and then backproject and fuse the results into 3D space. For example, SAI3D \cite{yin2024sai3d} introduces a hierarchical region-growing algorithm to refine 3D masks from SAM-generated \cite{kirillov2023segment} 2D segmentations, while Segment3D \cite{huang2024segment3d} uses the 2D results as pseudo ground-truth masks for 3D model pre-training. However, these approaches inherit challenges from 2D segmentation errors and multi-view inconsistencies, ultimately limiting their effectiveness. 

In this work, we propose a novel alternative that enhances generalization capabilities through cost-effective synthesis of 3D scenes with diverse object assets, simultaneously addressesing the data scarcity problem while circumventing the inherent limitations of 2D-based approaches.
\paragraph{3D Scene Synthesis}
Recently, 3D scene synthesis methods have made significant progress \cite{hu2024scenecraft,li2024dreamscene,ocal2024sceneteller,zhou2024dreamscene360}. For example, Holodeck \cite{yang2024holodeck} will prompt large language models (LLM) to select objects to be placed and design their positions, resulting in high visual quality and better alignment with user instructions. In contrast, RandomRooms \cite{rao2021randomrooms} adopts a random object selection and placement strategy, leading to diverse visual appearances. However, these methods are not suitable for our targeted task, as LLM will bias towards and repeat selecting most common objects so that violating geometry diversity and context complexity while random selection and placement will break layout reasonability. By contrast, we propose a new synthesis pipeline that enhances geometry diversity and context complexity while ensuring physically reasonable layouts, making the synthetic dataset better aligned with our task and leading to better model performance as shown in experiments.

\section{Methodology}
\label{headings}

\subsection{Problem Formulation and Learning Paradigm}
\label{sec:3.1}
Class-agnostic 3D instance segmentation addresses the task of segmenting all object instances in a 3D scene (typically represented as a point cloud) without identifying its semantic category. Formally, given a training dataset $\mathcal{D}=\{(\mathcal{P}_i, \hat{\mathcal{M}}_i)\}_{i=1}^N$ consisting of $N$ scenes, where $\mathcal{P}$ represents a 3D point cloud and $\hat{\mathcal{M}}$ denotes the set of ground-truth instance masks for all objects, we train a model $\Phi$ to predict a set of instance masks $\mathcal{M}$ for an input point cloud, supervised by the objective $\mathcal{L}$:
\begin{equation}
    \min_\Phi \mathcal{L} (\mathcal{M}, \hat{\mathcal{M}}) = \mathcal{L} (\Phi(\mathcal{P}), \hat{\mathcal{M}}).
\end{equation}
At inference time, $\Phi$ is expected to segment all object instances, including those unseen in $\mathcal{D}$, in arbitrary 3D scenes, posing significant generalization challenges.

Data diversity is a key factor to achieve generalization. However, real-world 3D datasets $\mathcal{D}^r$ often lack sufficient diversity due to the difficulties in 3D data acquisition and labeling. This limitation leads to suboptimal performance in methods trained exclusively on such constrained datasets \cite{takmaz2023openmask3d,nguyen2024open3dis}. Alternative approaches \cite{huang2024segment3d,yin2024sai3d}, which leverage 2D foundation models by segmenting objects in multi-view RGB-D images and lifting the results to 3D, face challenges like 2D segmentation errors and cross-view inconsistency. To address the data scarcity problem, we propose \textbf{ASSIST-3D}, an innovative 3D scene synthesis pipeline that procedurally synthesizes diverse, automatically annotated data $\mathcal{D}^s$. By combining $\mathcal{D}^r$ and $\mathcal{D}^s$, we formulate the following objective to enhance data diversity and improve generalization:
\begin{equation}
    \min_\Phi \mathcal{L} (\Phi(\mathcal{P}^r), \hat{\mathcal{M}^r}) + \alpha \mathcal{L} (\Phi(\mathcal{P}^s), \hat{\mathcal{M}^s}),
\end{equation}
where superscripts `r' and `s' denote real and synthetic sources, and $\alpha$ is a balancing parameter.

\subsection{ASSIST-3D: Adapted Scene Synthesis for Instance Segmentation in 3D}
\label{sec:3.2}


To generate the target synthetic 3D scene dataset $\mathcal{D}^s$, we propose \textbf{ASSIST-3D}, a purpose-built 3D scene generation pipeline for class-agnostic 3D instance segmentation.

\begin{figure*}[t]
\centering
\includegraphics[width=0.87\linewidth]{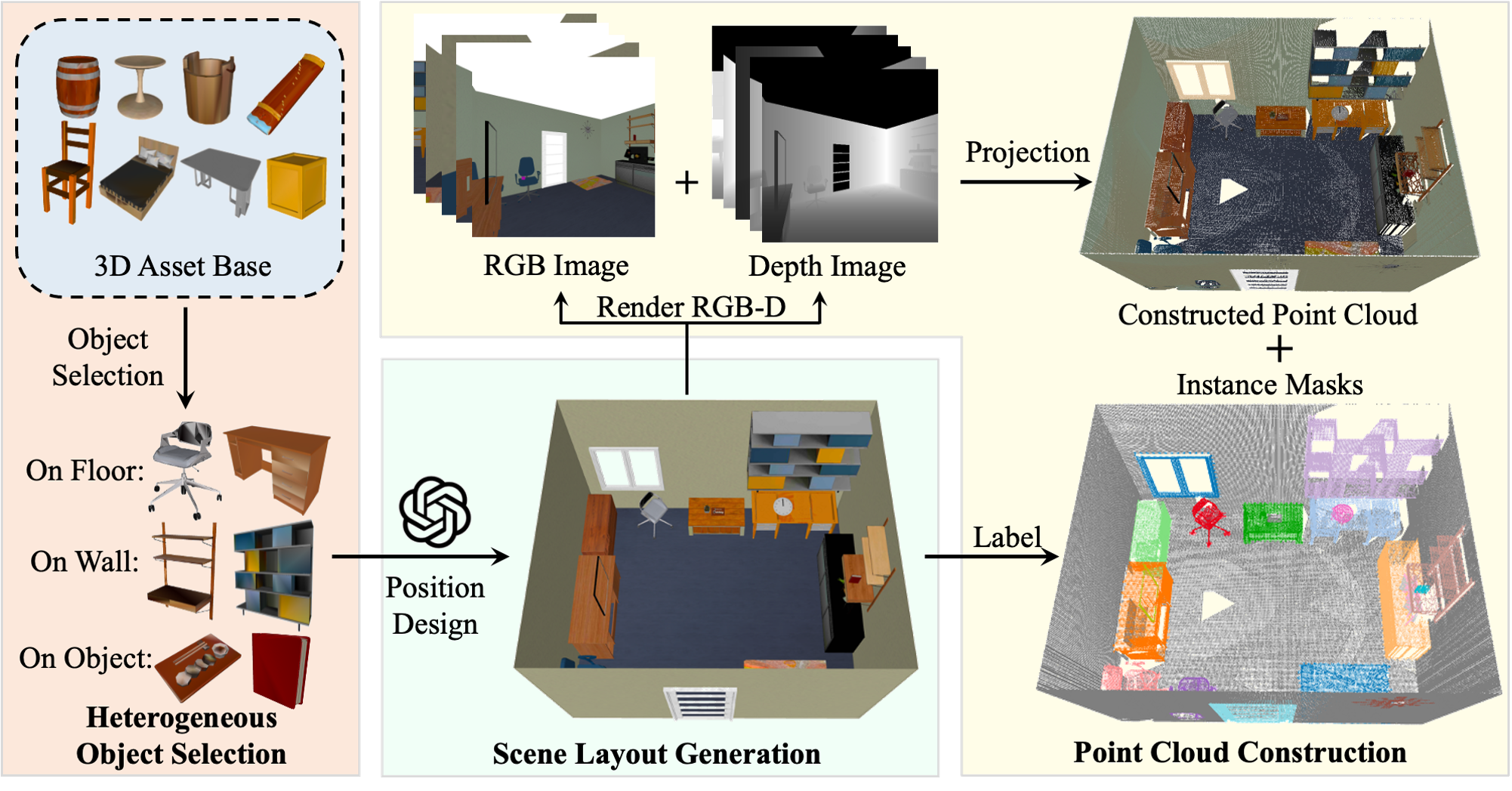}
\caption{An overview of our proposed ASSIST-3D. First, ASSIST-3D selects heterogeneous objects from a 3D asset base like Objaverse, satisfying geometry diversity and context complexity. Next, it leverages GPT-4 to design the arrangement of these objects within the scene, ensuring layout reasonability. Finally, it constructs realistic point clouds by mimicking the construction procedure of real dataset to reduce the domain gap and enhance performance.}
\label{img:2}
\end{figure*}

ASSIST-3D comprises three main stages, including \textbf{Heterogeneous Object Selection}, \textbf{Scene Layout Generation}, and \textbf{Realistic Point Cloud Construction}. Fig. \ref{img:2} gives a total illustration of the process.
First, ASSIST-3D utilizes a large collection of 3D CAD models as its asset base and selects heterogeneous objects by introducing randomness in selection of objects to be placed, ensuring rich variation in placed objects and contextual arrangements. 
Second, to generate physically plausible scenes, ASSIST-3D leverages LLM-based guidance, prompting GPT-4 to infer spatial relationships among objects and employing a depth-first search strategy for sequential placement. 
Finally, ASSIST-3D simulates real-world point cloud acquisition process by rendering multi-view RGB-D images of the synthetic scenes and merging the resulting partial point clouds. This process produces realistic point clouds along with instance mask annotations, together constituting the synthetic dataset.

\subsubsection{Heterogeneous Object Selection}
\label{sec:3.2.1}

To ensure geometry diversity, we begin with a large collection of 3D  models as our digital asset base, forming the foundation for data variety in our task. We select a subset of Objaverse \cite{deitke2023objaverse}, which contains 50,000 3D objects across 800 classes. Each object has been pre-processed to align its front view with the canonical coordinate axes. The asset is categorized into three groups, including 1) objects typically placed on the floor (e.g., furniture), 2) objects commonly mounted on or leaning against walls (e.g., photos, bookshelves), and 3) objects that can be placed on either of the first two groups. These groups are denoted as $\mathcal{O}_\text{floor}$, $\mathcal{O}_\text{wall}$, and $\mathcal{O}_\text{obj}$, respectively.

Leveraging the asset base, we can construct synthetic 3D scenes with high object diversity through simply randomized selection. Specifically, for each scene, we uniformly sample $M_1$ objects from $\mathcal{O}_\text{floor}$ and $M_2$ objects from $\mathcal{O}_\text{wall}$, denoted as $\bm{o}_\text{floor}$ and $\bm{o}_\text{wall}$, respectively. For each object in $\bm{o}_\text{floor}$ and $\bm{o}_\text{wall}$, we then sample 5 additional objects from $\mathcal{O}_\text{obj}$ to place on top of it, yielding a total of $5(M_1+M_2)$ objects, denoted as $\bm{o}_\text{obj}$. This sampling strategy effectively enhances the context complexity by disrupting conventional class co-occurrence patterns and increasing object diversity in one scene. To compensate for annotation incompleteness in real-world dataset $\mathcal{D}^r$, we implement a complementary sampling strategy that preferentially selects object categories present in $\mathcal{D}^r$ by adjusting their sampling probabilities. During scene synthesis, we alternate between these two strategies.

\subsubsection{Scene Layout Generation}
\label{sec:3.2.2}

With the selected objects, the next step is to decide their arrangements while ensuring reasonable scene layouts, bringing the synthetic scene closer to real-world environments. To achieve this target, we leverage a large language model (LLM) to infer physically plausible spatial relationships between objects and adopt a depth-first search (DFS) strategy for sequential object placement. 

More specifically, we first prompt GPT-4 to derive reasonable spatial relationships among objects in $\bm{o}_\text{floor}$, $\bm{o}_\text{wall}$, and each group within $\bm{o}_\text{obj}$. For structured placement, we prompt GPT-4 to sequentially determine object positions within each group, with the location of each object conditioned only on those placed earlier. Due to the limited spatial reasoning capabilities of GPT-4, its role is restricted to provide only object orientations and positions relative to previously objects, rather than absolute coordinates of objects in the scene. Our GPT-4 prompt template is provided in the appendix.

The sequential prompting enables our subsequent DFS-based placement strategy. Taking $\bm{o}_\text{floor}$ as an example, we begin by discretizing the floor space into uniform grids and initializing placement with the first object. At each iteration, we identify feasible grids for the current object, discarding those that violate spatial relationships provided by GPT-4. The object is then placed in one of the valid grids, and the process recurses with the next object. If no valid grid exists for the current object, the configuration of placement so far will be saved as a potential solution and backtrack to reposition the previous object. Among all solutions, we select the one that accommodates the most objects. 

The same procedure is applied to objects in $\bm{o}_\text{wall}$ and to each group within $\bm{o}_\text{obj}$, with placement occurring either on the wall or on supporting objects from $\bm{o}_\text{floor} \cup \bm{o}_\text{wall}$, as appropriate. Once the scene layout is generated, we place the objects accordingly to construct the final scene mesh.

\begin{table*}[ht]
\centering
\renewcommand{\arraystretch}{1.3}
\begin{tabular}{c|ccccccccc}
\shline
 \multirow{2}{*}{Method}   & \multicolumn{3}{c|}{ScanNet++}                              & \multicolumn{3}{c|}{S3DIS}                                     & \multicolumn{3}{c}{ScanNetV2}                     \\ \cline{2-10} 
                        & AP            & $\text{AP}_{50}$        & \multicolumn{1}{c|}{$\text{AP}_{25}$} & AP            & $\text{AP}_{50}$        & \multicolumn{1}{c|}{$\text{AP}_{25}$}        & AP            & $\text{AP}_{50}$        & $\text{AP}_{25}$        \\ \shline
Baseline                & 12.0          & 21.7          & \multicolumn{1}{c|}{32.7}   & 13.6          & 23.2          & \multicolumn{1}{c|}{41.9}          & 46.6          & 69.0          & 81.4          \\
Unscene3D~\cite{rozenberszki2024unscene3d}               & 9.8             & 18.6             & \multicolumn{1}{c|}{29.3}      & 21.4          & 40.3           & \multicolumn{1}{c|}{52.6}          & 15.9          & 32.2          & 58.5          \\
SAI3D~\cite{yin2024sai3d}                   & 17.1          & 31.1          & \multicolumn{1}{c|}{\textbf{49.5}}   & 24.8          & 42.4          & \multicolumn{1}{c|}{57.9}          & 30.8          & 50.5          & 70.6          \\
Segment3D~\cite{huang2024segment3d}               & 12.0          & 22.7          & \multicolumn{1}{c|}{37.8}   & 21.7          & 40.6          & \multicolumn{1}{c|}{53.4}          & 28.3         & 43.7          & 53.6          \\
Open3DIS~\cite{nguyen2024open3dis}                & 17.9          & 30.4          & \multicolumn{1}{c|}{39.7}   & 23.7          & 41.7          & \multicolumn{1}{c|}{56.3}          & 31.5          & 45.2          & 55.0          \\
SA3DIP~\cite{yangsa3dip}                  & 19.6          & 32.4          & \multicolumn{1}{c|}{42.5}   & 25.7          & 42.4          & \multicolumn{1}{c|}{58.6}          & 41.6          & 64.6          & 81.3          \\ \shline
\textbf{ASSIST-3D (Ours)}                    & \textbf{22.2} & \textbf{35.5} & \multicolumn{1}{c|}{45.2}   & \textbf{29.0} & \textbf{43.9} & \multicolumn{1}{c|}{\textbf{61.6}} & \textbf{48.1} & \textbf{70.7} & \textbf{83.4} \\ \shline
\end{tabular}
\caption{Performance comparison among different class-agnostic instance segmentation methods on ScanNet++ \cite{yeshwanth2023scannet++}, S3DIS \cite{armeni20163d}, and ScanNetV2 \cite{dai2017scannet}.}
\label{tab:1}
\end{table*}

\subsubsection{Realistic Point Cloud Construction}

\label{sec:3.2.3}

After generating the mesh of a synthetic 3D scene, a straightforward way to obtain its corresponding point cloud data is to sample points directly from the mesh surface. However, such point clouds tend to be overly uniform and lack the noise, density variations and occlusion commonly existing in real datasets. In practice, real-world point clouds are typically reconstructed via Simultaneous Localization and Mapping (SLAM) systems, which fuse multi-view RGB-D or LiDAR measurements from different vantage points in the scene, resulting in non-uniform sampling and occlusion. To bridge this domain gap, we propose a realistic point cloud construction method that closely mimics real-world acquisition processes. 

We first search for optimal scanning positions by uniformly partitioning the mid-height plane of the scene into $0.1\times 0.1 m^2$ regions while excluding areas occupied by objects. Using Farthest Point Sampling (FPS), we select five optimal vantage points for a comprehensive scene coverage. At each point, we render 12 RGB-D images by rotating camera in $30^{\circ}$ increments ($360^{\circ}$ total), yielding 60 images in total.

The subsequent process involves projecting depth maps into partial point clouds using camera intrinsics, then transforming and aggregating them into a unified global coordinate system according to camera poses. To further enhance realism, we apply voxel-based downsampling to discretize point cloud space and retains one random point per voxel. 

Finally, we generate semantic labels for the aggregated point cloud by assigning each point to its nearest object in the scene, producing a complete set of segmentation masks. Each point cloud paired with its corresponding mask set constitutes a training sample in the synthetic dataset.

\section{Experiments}
\label{sec:4}
\subsection{Experimental Settings}
\label{sec:4.1}
\noindent \textbf{Real-world Datasets} We conduct experiments on three widely used benchmarks, including ScanNetV2 \cite{dai2017scannet}, ScanNet++ \cite{yeshwanth2023scannet++}, and S3DIS \cite{armeni20163d}. ScanNetV2 provides dense instance masks for 18 categories, with official splits of 1,201 training and 312 validation scans. ScanNet++ enhances ScanNetV2 with increased point cloud number and labeled categories. S3DIS covers six indoor areas, with 13 semantic categories. We incorporate the training set of ScanNetV2 with ASSIST-3D-generated dataset for training and evaluate the model's in-domain performance on the validation set of ScanNetV2. To assess generalization, we test our method on ScanNet++ and S3DIS, which contain unseen object categories and follow different data distributions with ScanNetV2.

\noindent \textbf{Evaluation Metrics}  We follow \cite{yin2024sai3d, huang2024segment3d} to evaluate our method using the standard Average Precision (AP) metric. Specifically, we compute the mean AP score across Intersection-over-Union (IoU) thresholds ranging from $50\%$ to $95\%$ with $5\%$ increments (denoted as AP). Additionally, we report performance at specific IoU thresholds of $50\%$ and $25\%$ ($\text{AP}_{50}$ and $\text{AP}_{25}$ respectively). Consistent with the class-agnostic nature of our task, all reported metrics assess only instance mask accuracy, independent of semantic category recognition.

We further introduce metrics to quantify the synthetic data in accordance with the three proposed principles. To assess geometry diversity, we use 3DShape2VecSet \cite{zhang20233dshape2vecset} to extract the features of objects in the synthetic scenes. Since 3DShape2VecSet is a 3D VAE model trained for object reconstruction, its latent features effectively capture object geometry information. We compute the entropy of the resulting feature distribution, where higher entropy indicates greater geometry diversity.
To measure context complexity, given one class, we compute for each left class its occurrence probability in a scene containing the given class, and then average the maximal ones across all classes. Lower value suggests more complex and diverse contextual relationships among objects.
For assessing layout reasonability, we follow prior works \cite{yang2024semantic, yang2024llplace} by rendering multi-view images of each scene and prompting GPT-4o to score the layout on a scale from 0 to 100. A higher score indicates a more reasonable and realistic spatial arrangement. The exact prompt used for GPT-4o is provided in the appendix.


\noindent \textbf{Implementation Details} For each scene, the selected number of objects on floor and wall are $M_1=100$ and $M_2=50$. Using our proposed pipeline, we synthesize a dataset of 2,000 scenes with approximately 134,000 object instances in total, averaging 67 objects per scene. Half scenes are generated according to complementary strategy with 0.7 probability to select object categories in $\mathcal{D}^r$. We combine this synthetic dataset with ScanNetV2 to train Mask3D \cite{schult2023mask3d} by replacing its multi-class classifier with a binary objectness classifier (following \cite{huang2024segment3d}) and using binary cross-entropy loss, dice loss and mask loss as optimization objectives. Mask3D trained purely on ScanNetV2 will serve as the baseline. Training runs for 600 epochs with a batch size of 36 (distributed across 6 A100 GPUs). To mitigate domain gaps, we set the scaling factor $\alpha$ to 0.5 for optimization.

\begin{table*}[t]
	\centering
	\renewcommand{\arraystretch}{1.3}
    \resizebox{0.95\width}{!}{
	\begin{tabular}{c|ccc|cccccc}
		\shline

		\multirow{2}{*}{Method} &  \multirow{2}{*}{\begin{tabular}[c]{@{}c@{}}Geometry\\ Diversity\end{tabular}~($\uparrow$)} & \multirow{2}{*}{\begin{tabular}[c]{@{}c@{}}Context \\ Complexity\end{tabular}~($\downarrow$)} & \multirow{2}{*}{\begin{tabular}[c]{@{}c@{}}Layout\\ Reasonability\end{tabular}~($\uparrow$)} & \multicolumn{3}{c|}{ScanNet++}              & \multicolumn{3}{c}{S3DIS} \\ \cline{5-10} 
		                              &                   &                   &                   &AP  & AP$_{50}$ & \multicolumn{1}{c|}{AP$_{25}$} &   AP    &  AP$_{50}$     &   AP$_{25}$   \\ \shline

        Holodeck        &    $\times$~(0.85)                           &   $\times$~(0.38)                            &    \checkmark~(72)                   & 14.2 & 24.6   & \multicolumn{1}{c|}{36.3}   &    18.2     &   28.4     &    48.3    \\ 
        RandomRooms       &                \checkmark~(4.37)             &                   \checkmark~(0.04)           &              $\times$~(23)                 & 16.6 & 26.8   & \multicolumn{1}{c|}{38.4}   &    23.5     &    41.5    &    56.4    \\
		ASSIST-3D w/o Content        &       \checkmark~(4.03)                        &           $\times$~(0.54)                   &             \checkmark~(65)                & 19.4 & 31.9   & \multicolumn{1}{c|}{41.8}   &     26.4    &     42.8  &    58.6    \\ \shline
        ASSIST-3D (Ours)     &      \checkmark~(4.15)                         &          \checkmark~(0.08)                     &             \checkmark~(62)       & \textbf{22.2} & \textbf{35.5}   & \multicolumn{1}{c|}{\textbf{45.2}}   &    \textbf{29.0}     &    \textbf{43.9}    &   \textbf{61.6}     \\
        \shline
	\end{tabular}
    }
    \caption{Performance comparison among different 3D scene synthesis methods on ScanNet++ \cite{yeshwanth2023scannet++}  and S3DIS \cite{armeni20163d}. Evaluation metrics of geometry diversity, context complexity, and layout reasonability are feature entropy, object co-occurrence probability, GPT-4o prompted layout score, respectively, as detailed in Sec. \ref{sec:4.1}.}
	\label{tab:2.1}
\end{table*}

\subsection{Comparison with State-of-The-Arts}
\label{sec:4.2}
We first compare the performance of our ASSIST-3D method with other state-of-the-art methods, including Open3DIS \cite{nguyen2024open3dis}, Unscene3D \cite{rozenberszki2024unscene3d}, Segment3D \cite{huang2024segment3d}, SAI3D \cite{yin2024sai3d}, and SA3DIP \cite{yangsa3dip}. Among these methods, SAI3D and SA3D leverage SAM \cite{kirillov2023segment} to generate 2D masks from multi-view RGB-D images, then refine 3D masks through projection and aggregation of these 2D predictions. Other methods build upon Mask3D, trained on ScanNetV2 using either limited manual annotations (e.g., Open3DIS) or pseudo-3D masks derived from SAM-generated 2D masks (e.g., Unscene3D and Segment3D). For a fair comparison, we also train a version of Mask3D on a combination of annotated ScanNetV2 data and our synthetic dataset, and establish a baseline using only the original ScanNetV2 annotations. 

As shown in Table \ref{tab:1}, our method achieves the best performance across all the benchmarks. 
Notably, it enables Mask3D to retain strong accuracy on the in-domain ScanNetV2, where other methods struggle due to noisy 2D masks and limited 3D annotations. While existing methods exhibit decent generalization and outperform the baseline on ScanNet++ and S3DIS, our approach further enhances Mask3D with synthetic data generated by our proposed ASSIST-3D pipeline.

We also compare ASSIST-3D with other 3D scene synthesis methods, including Holodeck \cite{yang2024holodeck}, which differs in selecting objects for placement by leveraging GPT-4 and thus favors common objects and typical co-occurrence patterns, and RandomRooms \cite{rao2021randomrooms}, which relies on completely random object selection and placement.  As shown in Table~\ref{tab:2.1}, synthetic data generated by Holodeck lacks geometry diversity and context complexity, while that generated by RandomRooms suffers from poor layout reasonability. To isolate the effects of geometry diversity and context complexity, we further introduce a variant of ASSIST-3D that disables context complexity. Specifically, each object class in the 3D asset base is paired with a co-occurring class. The paired class will also be selected with 66\% probability when selecting objects from one class for placement. The performance gains achieved by the full ASSIST-3D pipeline on ScanNet++ and S3DIS underscore the importance of incorporating all three principles in generating high-quality synthetic data for class-agnostic 3D instance segmentation.


\subsection{Ablation Studies}
\label{sec:4.2.2}

\begin{table}[t]
\centering
\renewcommand{\arraystretch}{1.3}
\begin{tabular}{c|cccccc}
\shline
\multirow{2}{*}{\begin{tabular}[c]{@{}c@{}}Cluster \\ Number\end{tabular}} & \multicolumn{3}{c|}{ScanNet++}                                       & \multicolumn{3}{c}{S3DIS}                   \\ \cline{2-7} 
& AP   & AP$_{50}$ & \multicolumn{1}{c|}{AP$_{25}$} & AP & AP$_{50}$ & AP$_{25}$ \\ \shline
1                            & 14.6 & 24.7   & \multicolumn{1}{c|}{36.1}     &  16.7  &    28.2    &    47.2    \\
2                            & 16.8 & 27.3   & \multicolumn{1}{c|}{38.9}  &  21.1  &    34.7    &    52.3    \\
3                            & 17.8 & 29.6   & \multicolumn{1}{c|}{40.3}   &  24.1  &    39.7    &    56.4     \\
4                            & 20.1 & 32.4   & \multicolumn{1}{c|}{43.2}   &  27.3  &    41.6    &    58.7     \\
5 (Ours)                     & \textbf{22.2} & \textbf{35.5}   & \multicolumn{1}{c|}{\textbf{45.2}} &  \textbf{29.0}  &    \textbf{43.9}    &    \textbf{61.6}      \\ \shline
\end{tabular}
\caption{Performance comparison of ASSIST-3D with different numbers of object clusters for data synthesis.}
\label{tab:3}
\end{table}

\begin{table}[t]
\centering
\renewcommand{\arraystretch}{1.3}
\resizebox{0.96\width}{!}{
\begin{tabular}{c|cccccc}
\shline
\multirow{2}{*}{\begin{tabular}[c]{@{}c@{}}Cluster \\ Number\end{tabular}} & \multicolumn{3}{c|}{ScanNet++}                                       & \multicolumn{3}{c}{S3DIS}                   \\ \cline{2-7} 
& AR   & AR$_{50}$ & \multicolumn{1}{c|}{AR$_{25}$} & AR & AR$_{50}$ & AR$_{25}$ \\ \shline
1                              & 3.2  & 6.7    & \multicolumn{1}{c|}{8.4}        &  13.3  &    23.8    &    35.2    \\
2                              & 6.7  & 11.3   & \multicolumn{1}{c|}{15.2}     &  16.1  &    28.1    &    39.2    \\
3                              & 8.6  & 15.2   & \multicolumn{1}{c|}{20.9}       &  18.6  &   32.4     &    44.6    \\
4                              & 10.6 & 18.4   & \multicolumn{1}{c|}{25.3}       &  21.2  &   36.4     &    48.0    \\
5 (Ours)                       & \textbf{12.4} & \textbf{21.6}   & \multicolumn{1}{c|}{\textbf{29.8}}     &  \textbf{23.6}  &    \textbf{40.2}    &   \textbf{51.8}     \\ \shline
\end{tabular}
}
\caption{Performance comparison of ASSIST-3D using different numbers of object clusters for data synthesis on test objects with geometries distinct from those in ScanNetV2 \cite{dai2017scannet}.}
\label{tab:3.1}
\end{table}

\begin{figure*}[t]
\centering
\includegraphics[width=0.8\linewidth]{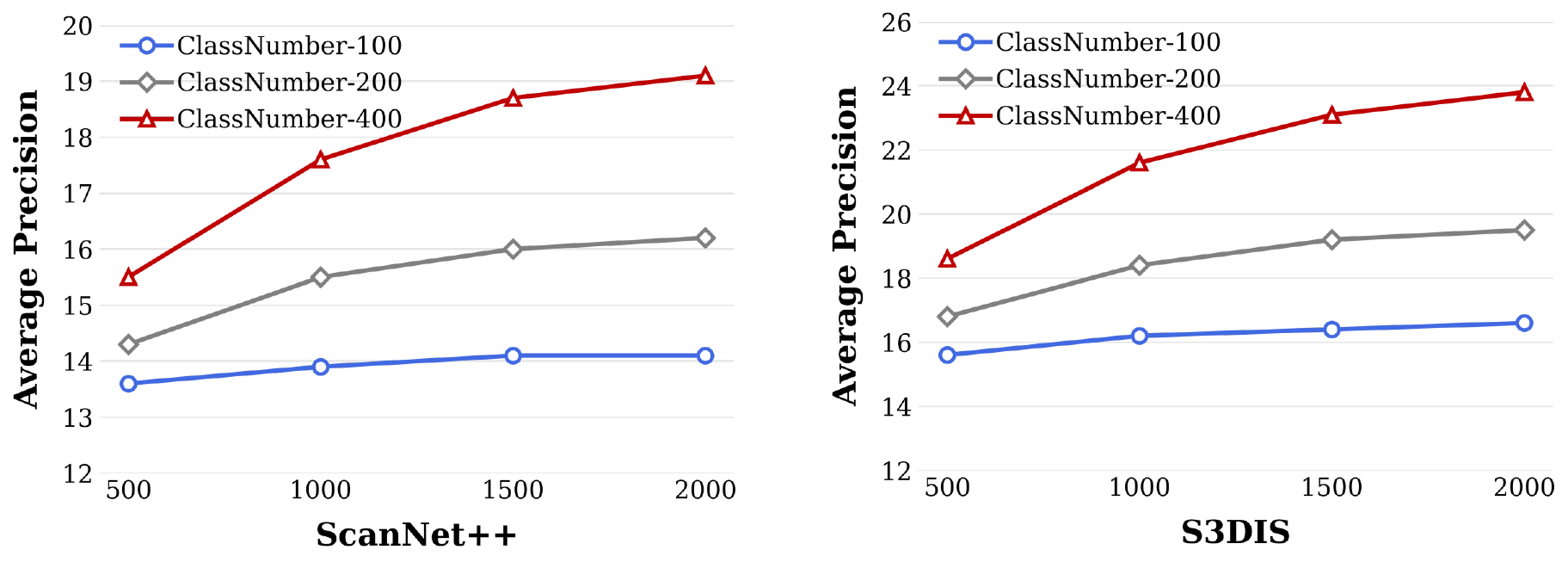}
\caption{Performance curves of average precision on ScanNet++ \cite{yeshwanth2023scannet++} and S3DIS \cite{armeni20163d} with varying numbers of object classes used for synthesis and different amounts of synthetic training data.}
\label{img:3}
\end{figure*}

\paragraph{Effect of Geometry Diversity} 
In Section~\ref{sec:4.2}, we have demonstrated the importance of geometry diversity in synthetic data generation. Here, we further validate its impact on model performance through controlled experiments. We begin by extracting features for all 3D assets using 3DShape2VecSet \cite{zhang20233dshape2vecset}, and apply the k-means clustering algorithm to group them into five distinct clusters based on geometric similarity. To vary geometry diversity, we synthesize training datasets using an increasing number of clusters to select objects from. As in Table~\ref{tab:3}, model performance consistently improves with greater geometry diversity.

To further assess the effectiveness of geometry diversity, we evaluate performance on out-of-domain objects, specifically, those in ScanNet++ and S3DIS that are geometrically distinct from those in ScanNetV2. We extract features for all objects in ScanNetV2 using 3DShape2VecSet and compute the mean feature vector $\mathcal{F}_{avg}$. We then calculate the maximum L2 distance $d_{max}$ from $\mathcal{F}_{avg}$ to any object feature in ScanNetV2. For each object in ScanNet++ or S3DIS, we extract its feature $\mathcal{F}$ and compute its L2 distance $d$ to $\mathcal{F}_{avg}$. An object is considered out-of-domain and selected for evaluation if $d > d_{max}$. We report the mean Average Recall (AR) over IoU thresholds from 50\% to 95\% (in 5\% increments), as well as $\text{AR}_{50}$ and $\text{AR}_{25}$, which represent recall at IoU thresholds of 50\% and 25\%, respectively. The results, presented in Table~\ref{tab:3.1}, are consistent with those in Table~\ref{tab:3}, further confirming the effectiveness of geometry diversity in enhancing generalization to novel object geometries.

\paragraph{Effect of Context Complexity}

In this part, we design experiments to quantify the effect of context complexity. We begin by pairing each class in our 3D asset base with a co-occurring class. For classes present in ScanNetV2, we pair them with their most frequently co-occurring counterparts based on the ScanNetV2 training data, while the remaining classes are paired randomly. To control context complexity and co-occurrence probability, we define four levels by varying the probability of simultaneously selecting from paired class when selecting objects from one class for placement: 100\%, 66\%, 33\%, and 0\% (i.e., no co-occurrence enforcement). The corresponding object co-occurrence probabilities and performance results are reported in Table~\ref{tab:4}, which show that increasing context complexity consistently leads to improved performance on both ScanNet++ and S3DIS.


\begin{table}[t]
	\centering
	\renewcommand{\arraystretch}{1.3}
    \resizebox{0.87\width}{!}{
	\begin{tabular}{c|c|cccccc}
		\shline
		\multirow{2}{*}{Level} &\multirow{2}{*}{Prob} & \multicolumn{3}{c|}{ScanNet++}                              & \multicolumn{3}{c}{S3DIS}                                                     \\ \cline{3-8} 
		& & AP            & $\text{AP}_{50}$        & \multicolumn{1}{c|}{$\text{AP}_{25}$} & AP            & $\text{AP}_{50}$        & \multicolumn{1}{c}{$\text{AP}_{25}$}        \\ \shline
		1          &  100\%    & 17.2          & 29.4          & \multicolumn{1}{c|}{36.3}   & 24.3          & 42.0          & \multicolumn{1}{c}{57.6}          \\
		2          &   66\%  &     19.4      &    31.9       & \multicolumn{1}{c|}{41.8}   & 26.4         & 42.8          & \multicolumn{1}{c}{58.6}          \\
		3         &   33\%   & 21.6          & 34.7          & \multicolumn{1}{c|}{44.6}   & 28.2          & 43.4          & \multicolumn{1}{c}{60.7}          \\
		4 (Ours)       &     0\%        & \textbf{22.2} & \textbf{35.5} & \multicolumn{1}{c|}{\textbf{45.2}}   & \textbf{29.0} & \textbf{43.9} & \multicolumn{1}{c}{\textbf{61.6}} \\ \shline
	\end{tabular}
    }
\caption{Performance comparison of ASSIST-3D with different levels of context complexity.`Prob' denotes the probability of simultaneously sampling objects from paired class of one class for placement to control context complexity.}
\label{tab:4}
\end{table}
\begin{table}[t]
\centering
\renewcommand{\arraystretch}{1.3}
\resizebox{0.86\width}{!}{
\begin{tabular}{c|cccccc}
\shline
 \multirow{2}{*}{\begin{tabular}[c]{@{}c@{}}Point Cloud \\ Construction\end{tabular}}                       & \multicolumn{3}{c|}{ScanNet++}                              & \multicolumn{3}{c}{S3DIS}                                    \\ \cline{2-7} 
                        & AP            & $\text{AP}_{50}$        & \multicolumn{1}{c|}{$\text{AP}_{25}$} & AP            & $\text{AP}_{50}$        & \multicolumn{1}{c}{$\text{AP}_{25}$}                \\ \shline
-               & 12.0          & 21.7          & \multicolumn{1}{c|}{32.7}   & 13.6          & 23.2          & \multicolumn{1}{c}{41.9}                    \\
Mesh-sampled                  & 14.2          & 24.6          & \multicolumn{1}{c|}{36.3}   & 18.8          & 30.5        & \multicolumn{1}{c}{51.5}                   \\
Rendering-based                    & \textbf{22.2} & \textbf{35.5} & \multicolumn{1}{c|}{\textbf{45.2}}   & \textbf{29.0} & \textbf{43.9} & \multicolumn{1}{c}{\textbf{61.6}} \\ \shline
\end{tabular}
}
\caption{Performance comparison of various strategies of point cloud construction in ASSIST-3D.}
\label{tab:5}
\end{table}

\paragraph{Effect of Realistic Point Cloud Construction}

As described in Section~\ref{sec:3.2.3}, we replicate the point cloud construction process commonly used in real-world datasets. This involves rendering RGB-D images from multiple viewpoints within each synthetic scene, projecting the depth pixels into 3D space, and applying voxel-based downsampling. This pipeline helps bridge the domain gap between synthetic and real data. To evaluate its effectiveness, we compare it against a baseline where point clouds are directly sampled from the synthetic meshes.
As shown in Table~\ref{tab:5}, Mask3D trained on mesh-sampled point clouds offers limited improvement over the baseline without synthetic data, whereas our rendering-based approach yields significantly better performance. 



\paragraph{Scalability of Synthetic Data}
In this part, we analyze the scalability of the synthetic data in terms of both the number of classes and the number of scenes. Specifically, we consider three levels for the number of classes used in scene synthesis: 100, 200, and 400; and four levels for the number of synthesized scenes: 500, 1000, 1500, and 2000. For each combination of class levels and scene levels, we generate a synthetic dataset and combine it with ScanNetV2 to train the adapted Mask3D. As illustrated in Fig.~\ref{img:3}, the AP performance on both ScanNet++ and S3DIS consistently improves with an increasing number of classes. Furthermore, the benefit of synthesizing more scenes becomes more pronounced when the number of object classes is higher. These results demonstrate that scaling both the class diversity and dataset size enhances the model's generalization ability.


\section{Conclusion}
In this paper, we propose ASSIST-3D for class-agnostic 3D instance segmentation. To address the challenges of limited annotations in real dataset and noisy 2D segmentations, we propose to synthesize 3D scene-level dataset for training. Rather than replicating previous 3D scene synthesis methods, we design a new process that simultaneously satisfies the principle of geometry diversity, context complexity and layout reasonability. Additionally, we mimic the point cloud construction of real datasets to reduce the domain gap between real and synthetic dataset. By training Mask3D on ScanNetV2 augmented with our synthetic dataset, it achieves the best performance on ScanNet++, S3DIS, and ScanNetV2.

\section*{Acknowledgments}
The work has been supported by Hong Kong Research Grant Council - Early Career Scheme (Grant No. 27209621), General Research Fund Scheme (Grant No. 17202422, 17212923, 17215025), Theme-based Research (Grant No. T45-701/22-R) and Shenzhen Science and Technology Innovation Commission (SGDX20220530111405040).  Part of the described research work is conducted in the JC STEM Lab of Robotics for Soft Materials funded by The Hong Kong Jockey Club Charities Trust.

\bibliography{aaai2026}

\end{document}